\documentclass{isprs} 
\usepackage{subfigure}
\usepackage{setspace}
\usepackage{geometry} 
\usepackage{epstopdf}
\usepackage[labelsep=period]{caption}  
\usepackage[british]{babel} 
\usepackage[hang]{footmisc}

\begin{document}

\title{Building Change Detection using Multi-Temporal Airborne LiDAR Data}
\date{}

\author{Ritu Yadav, Andrea Nascetti, Yifang Ban\thanks{Corresponding author}}


\address{Division of Geoinformatics, KTH Royal Institute of Technology, Sweden - (rituy, nascetti, yifang)@kth.se}


\icwg{}   

\abstract{
Building change detection is essential for monitoring urbanization, disaster assessment, urban planning and frequently updating the maps. 3D structure information from airborne light detection and ranging (LiDAR) is very effective for detecting urban changes. But the 3D point cloud from airborne LiDAR(ALS) holds an enormous amount of unordered and irregularly sparse information.  Handling such data is tricky and consumes large memory for processing. Most of this information is not necessary when we are looking for a particular type of urban change. In this study, we propose an automatic method that reduces the 3D point clouds into a much smaller representation without losing necessary information required for detecting Building changes. The method utilizes the Deep Learning(DL) model U-Net for segmenting the buildings from the background. Produced segmentation maps are then processed further for detecting changes and the results are refined using morphological methods. For the change detection task, we used multi-temporal airborne LiDAR data. The data is acquired over Stockholm in the years 2017 and 2019. The changes in buildings are classified into four types: ‘newly built’, ‘demolished’, ‘taller’ and 'shorter'. The detected changes are visualized in one map for better interpretation. 
}


\maketitle

\section{Introduction}
\label{sec:intro}

With the population growth and rapid development, it became crucial to monitor uncontrolled urbanization. The rate of people moving from villages to cities is increasing at a threatening level. Urban changes can be due to the construction of new railway tracks, roads, construction of the new residential area, demolition of old or illegal properties, filling lakes, and many others. For sustainable urban planning, regular updating of footprints and maps is becoming challenging every year. In such a scenario, remote change detection can play a significant role in monitoring and understanding the local and global changes. Change detection can also facilitate environmental monitoring, resource exploration, and disaster assessment.
Conventionally, the urban changes were detected using methods developed on multi-temporal remote sensed multi-spectral or optical images. The high-resolution optical data is useful in change detection but is dependent on sunlight, fog, and multiple other environmental conditions. On the other hand, LiDAR and Radar active remote sensors are immune to these factors and better at detecting urban changes. Radar data is difficult to process and involves a lot of noise issues. In this scenario, LiDAR data is an effective alternative. LiDAR provides 3D data, which helps in monitoring volumetric changes as well as 2D changes. ALS has dramatically changed the monitoring and 3D modeling of districts, cities, etc. LiDAR provides information such as elevation, intensity, and 'number of returns', which helps in segregating different features in the scene. These attributes are significant in detecting buildings. 

In previous studies, ALS data has shown great potential in detecting changes in buildings. The methods are further explored in combination with aerial images. One example of such work is ~\cite{matikainen2010automatic}, where a combination of aerial and ALS data is used to detec the buildings. For detecting buildings, elevation information from ALS is used and arial images are used to segregate trees from buildings. Changes in buildings are detected using object based matching algorithm.  In ~\cite{murakami1999change} study, multi-temporal high resolution DSM is prepared from the ALS data and changes are detected by calculating difference between the DSM. Authors of study ~\cite{teo2013lidar}, also explored the multi-temporal DSM prepared from ALS data.  DSM are first classified on the basis of surface roughness and the changes in buildings is detected by change in the land cover class. In ~\cite{vu2004lidar} study, authors proposed a building change detection method on LiDAR data in highly dense urban area. Here high resolution ALS grid data is used instead of point data. Building changes are detected by manually thresholding the difference histogram of the two grids. In ~\cite{dos2020automatic} study also authors subtract two DSM from the two time stamps and refined the changes using a height entropy concept.

\begin{figure*}  
\centering  
\begin{subfigure}
  \centering  
  \includegraphics[width=0.90\textwidth, height=105mm]{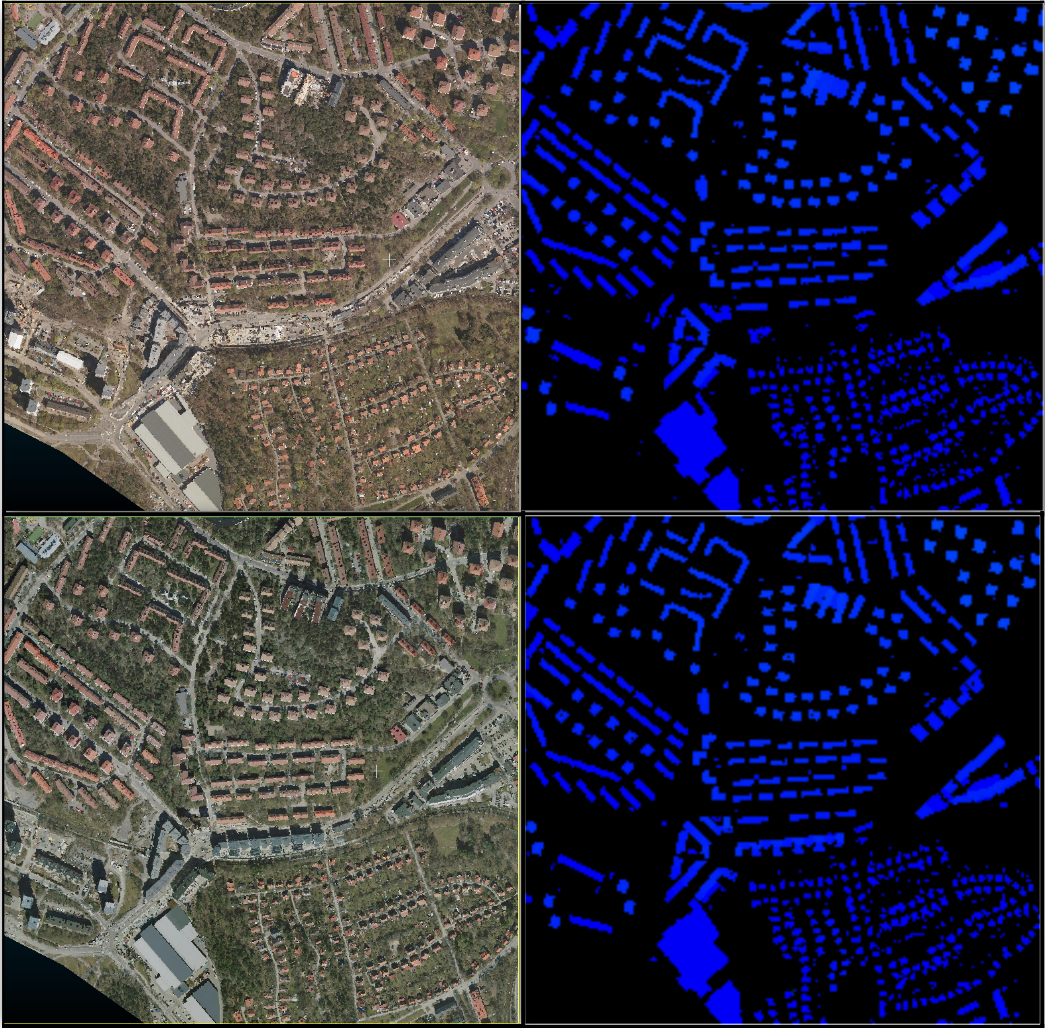}
\end{subfigure}
\caption{Data Sample Visualization. From left to right, we have RGB image which is the orthophoto projection on LiDAR point cloud, second is Ground truth used for the Building Segmentation task. First row is showing data from year 2017 and second from 2019.}
\label{fig1}
\end{figure*}

This work is conducted on multi-temporal ALS data to detect urban changes over the Stockholm area. In the proposed change detection method we first use DL segmentation methods to detect all buildings and then proceed with the change detection process. Pointnet~\cite{qi2017pointnet} and pointnet++~\cite{qi2017pointnet++} are the two recent segmentation methods popular for there state of the art performance on 3D point cloud data. These DL methods work on consistent LiDAR data but ALS suffers from the problem of random irregular data issues. In our ALS dataset, For the majority of sites, LiDAR data points are available only on the building rooftop. Since, 3D structure information is missing randomly across the data, using 3D point-cloud segmentation methods will cause false change detection in the later process. With this knowledge, we decided to drop the idea of using 3D point cloud segmentation and figure out a way to segment building in 2D keeping all important features intact.

In this study, we propose one single stream segmentation network which works on selected LiDAR attributes. The same network can be trained on a segmentation task for sparse optical data associated with LiDAR data points. We also presented one dual-stream segmentation architecture where selected LiDAR features and optical features are fused to get more accurate segmentation predictions.
Before calculating changes between the two-year buildings, we complemented our segmentation results with interpolation and morphology-based methods. The presented method attempts to perform the task in multiple steps, which are namely data processing, 2D projection, segmentation, building map preparation, area and elevation-wise change detection of buildings, refining the change maps using morphological algorithms, final change map visualization, and qualitative assessment of the results.

\section{Study Area and Data}
\label{sec:Dataset}

In this study, the dataset used is recorded over multiple sites in Stockholm during April 2017 and April 2019. The data is private and belongs to the Stockholm city government. The dataset is 3D point cloud data recorded using LiDAR airborne scanners with the point density of 12 points per meter square ground area. 

In total, 27 different sites are scanned and for each site, we have two points clouds from the years 2017 and 2019. Every point cloud files contain 6 million to 35 million points. Each point is associated with 12 different information, such as point coordinates, RGB values, intensity, elevation, number of returns, scan angle rank, GPS time, and point source id. For the change detection task explained in the following section, we extracted the point coordinates and corresponding six attribute values R, G, B, elevation, intensity, and 'number of returns'. In the dataset, Building size varies from 60 meter square to 1700 meter square. Also, Ground surfaces are relatively continuous and flat. For a better understanding of the area, one sample from both 2017 and 2019 is visualized in Figure\ref{fig1}. 

There are few black sites in the dataset, where data is not available. In multiple areas, LiDAR data points are sparse and missing. RGB color information is collected by projecting the Orthophoto of each area on to corresponding 3D point cloud. Hence the RGB information in the dataset is also affected by the black thin line and other Lidar point cloud issues. The most frequent case is missing data on top of the building, creating small holes. These data points are missing inconsistently between the two-year data and cause false change detection. Since, the changes are really small in area, to a large extent we handled this using interpolation. 

\section{Method}
\label{sec:seg_method}
\begin{figure}[htbp]
\centerline{\includegraphics[width=\columnwidth, height=115mm]{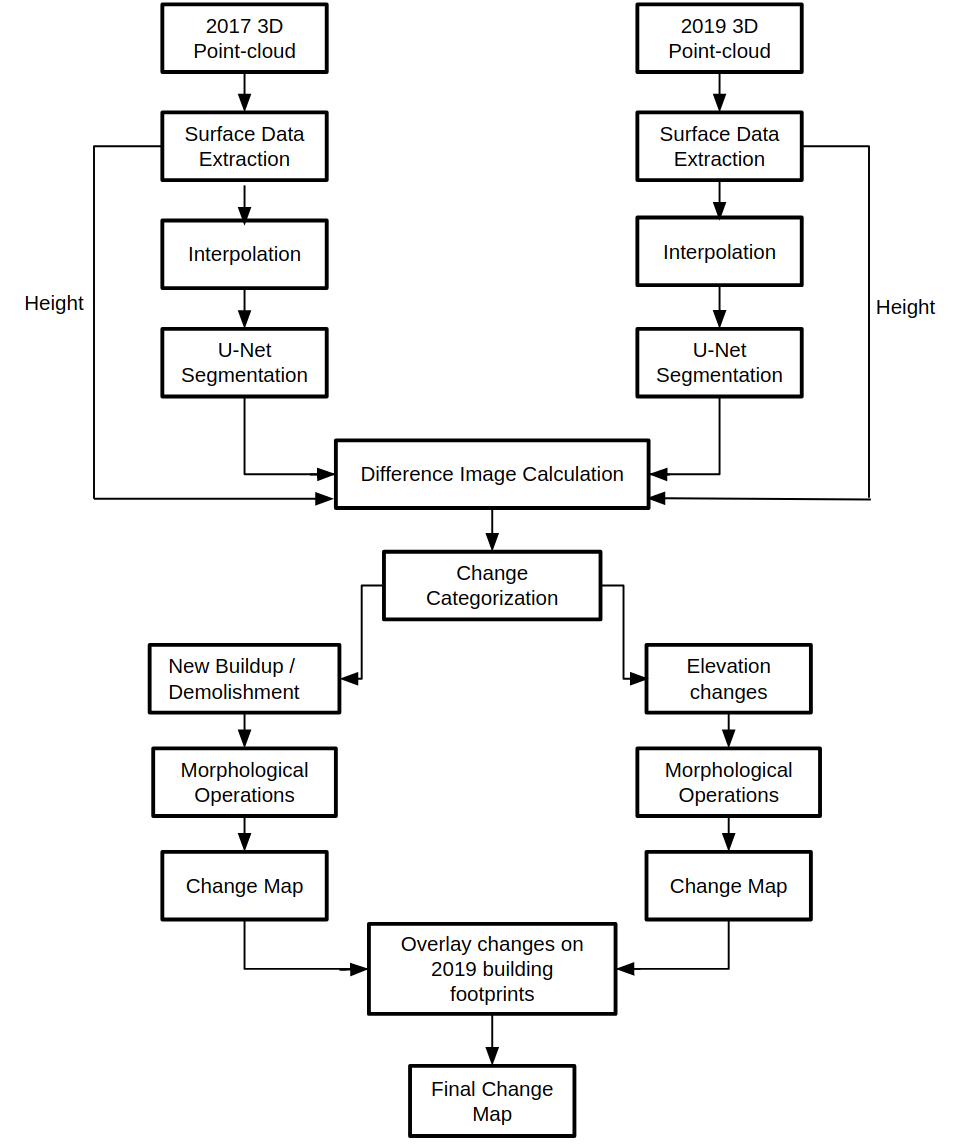}}
\caption{Building Change Detection Workflow.}
\label{fig2}
\end{figure}

The complete workflow of the building change detection is illustrated in Figure \ref{fig2} and the steps involved are explained in the following subsections.

\subsection{Pre-Processing}
\label{preprocess_method}
Handling a 3D point cloud containing millions of points is complicated to process as well as heavy on memory. Due to random order of missing 3D building structure information, it is difficult to maintain the data consistency between the two years. Detecting changes using inconsistent data will cause false detection. For detecting building changes, it is not necessary to utilize all information associated with each point in the point cloud. The idea is to extract necessary information from the 3D point cloud and then perform the change detection task. Since we are interested in detecting spatial changes and the changes in elevations, the top view from the point cloud is sufficient. For the top view, we need to gather the surface data from the point cloud, which is done by extracting the surface point for each pixel. The surface point at a given location is the data point with the maximum elevation value. The extracted surface data is then projected on a 2D plane.

From the data, a 3-channel matrix is prepared, where each channel is the 2D projection of the 3D point cloud. The first channel is holding the 'elevation' information. The second and third channel contains 'intensity' and 'number of returns' data respectively. Both 2017 and 2019 point cloud is processed in the same way. Similarly, we extract the RGB values of the 2D projected points and place them in 3 channels.

\subsection{Single Stream Segmentation Network}
\label{sec:seg_method}
For the segmentation task, lightweight U-net~\cite{ronneberger2015u} architecture is proposed. U-Net is one of the widely used CNN architectures originally designed for the pixel-level classification of medical images. This network consists of an encoder-decoder architecture. The encoder extracts the salient features from the input using several convolutional blocks, max-pooling, and batch normalization. The output of the encoder is a smaller set of feature maps. These small features are decoded back to the bigger feature maps guided to have a limited number of classes. The output is a segmentation map, where each pixel is classified.

Conventionally, Resnet~\cite{he2016deep} is the encoder in U-net architecture. In our proposed network, we replaced resnet backbone with a much smaller network known as Efficientnet-B4~\cite{tan2019efficientnet}. In addition to this, we only used 3 levels of the encoder, unlike most U-net architectures. In the decoder network, the encoded feature maps are up-sampled using a series of transpose convolution, convolutional layer, and batch normalization layers. Overall our proposed network settings contain one-fourth of the parameter in comparison to resnet based U-net, which makes it significantly lightweight.

\begin{figure}[htbp]
\centerline{\includegraphics[height=30mm, width=\columnwidth]{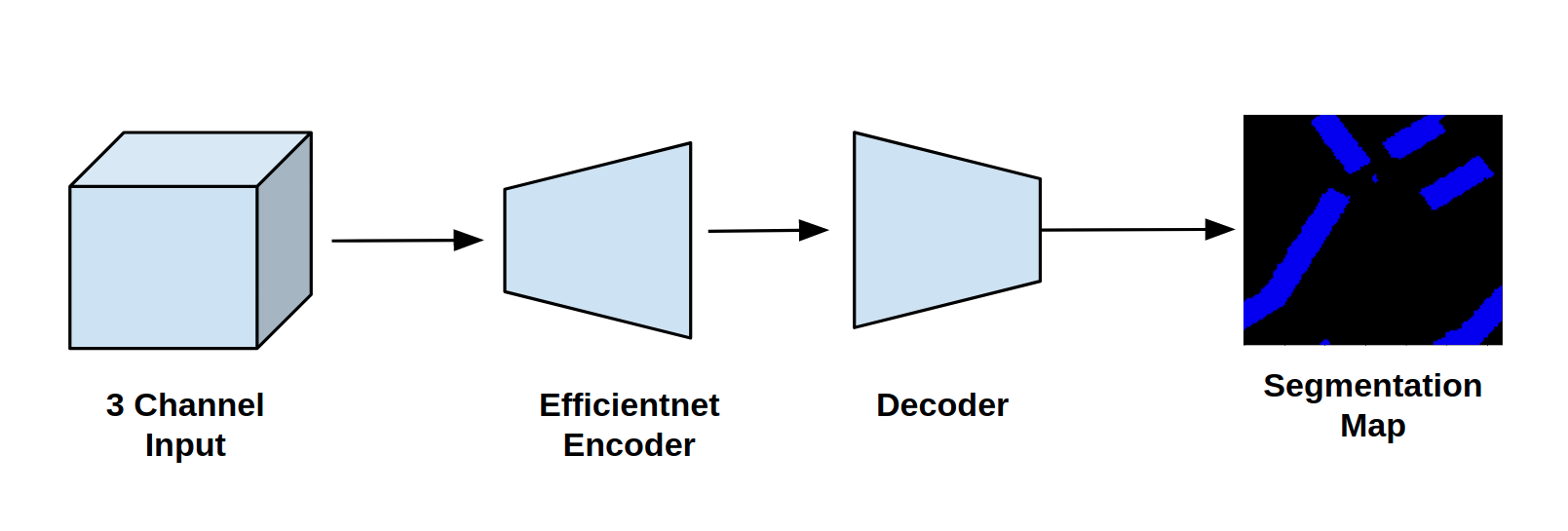}}
\caption{Single Stream Segmentation Network.}
\label{fig5}
\end{figure}

The network takes The 3 channel input(ZIN) containing elevation(Z), intensity(I), and the number of returns(N) information. The output of the network is a segmentation map with two classes named building and background. This is a single stream lightweight U-net architecture. For better understanding, the network is represented using a block diagram in Figure \ref{fig5}.

\subsection{Dual Stream Fusion Segmentation Network}
\label{sec:seg_method}
\begin{figure}[htbp]
\centerline{\includegraphics[height=45mm, width=0.51\textwidth]{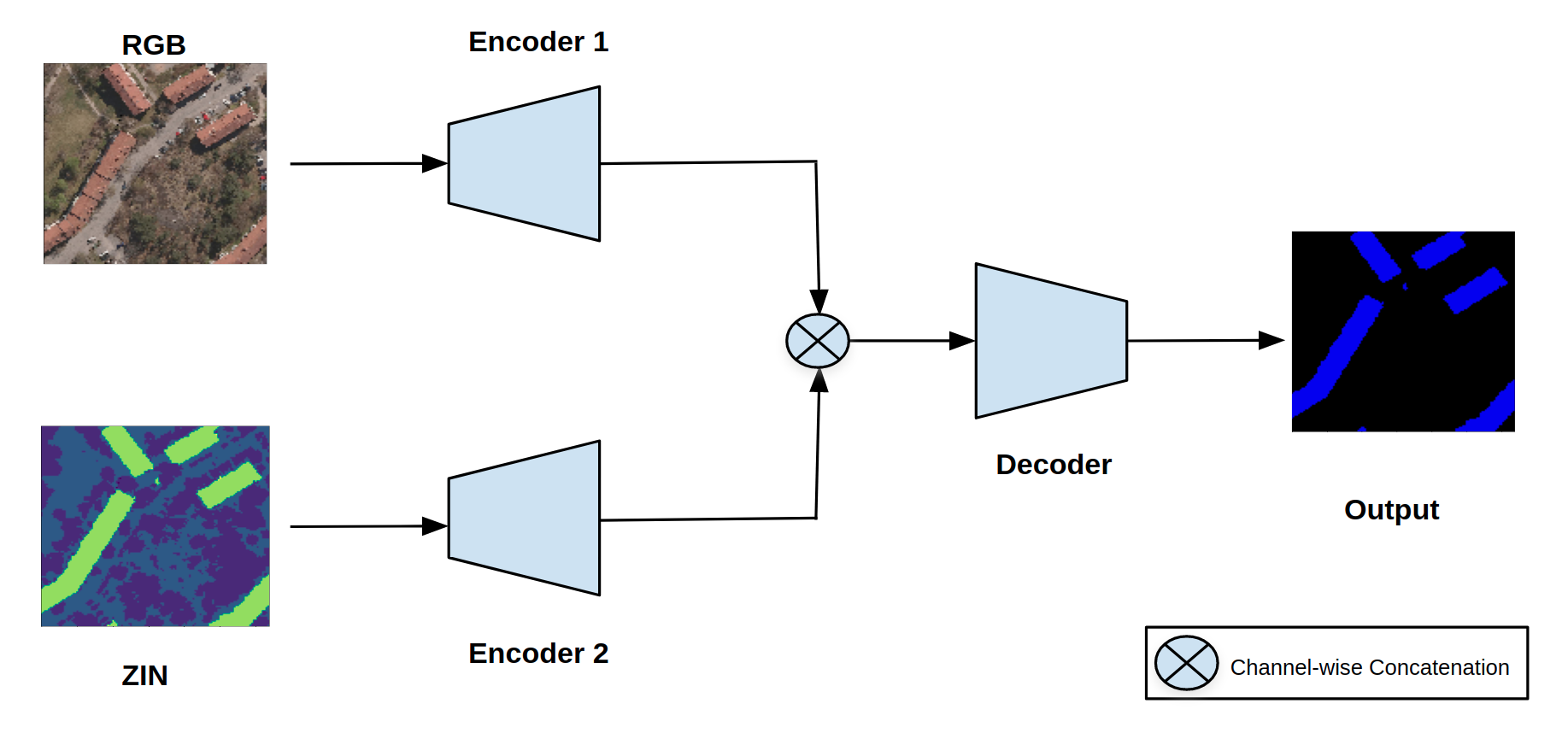}}
\caption{RGB and LiDAR Fused Dual Stream Segmentation Network.}
\label{fig6}
\end{figure}

As mentioned in subsection\ref{preprocess_method}, we have the RGB values corresponding to all the LiDAR data points. We wanted to compliment our network with optical information. To achieve that, we propose a dual-stream segmentation network where the network takes two 3-channel input streams, one with RGB values and the second(ZIN) stacking elevation, intensity, and the number of returns. The inputs from the two streams are encoded separately using two Efficientnet-B4 encoders. The encoded features are then fused and supplied to the decoder network to prepare a binary segmentation map. The encodes feature maps from the two streams are fused using a concatenation operation. The block diagram of our dual-stream segmentation network is presented in Figure \ref{fig6}.

\begin{figure*}  
\centering  
\begin{subfigure}
  \centering  
  \includegraphics[trim=1cm 0.2cm 1cm 1cm, width=\textwidth, height=65mm]{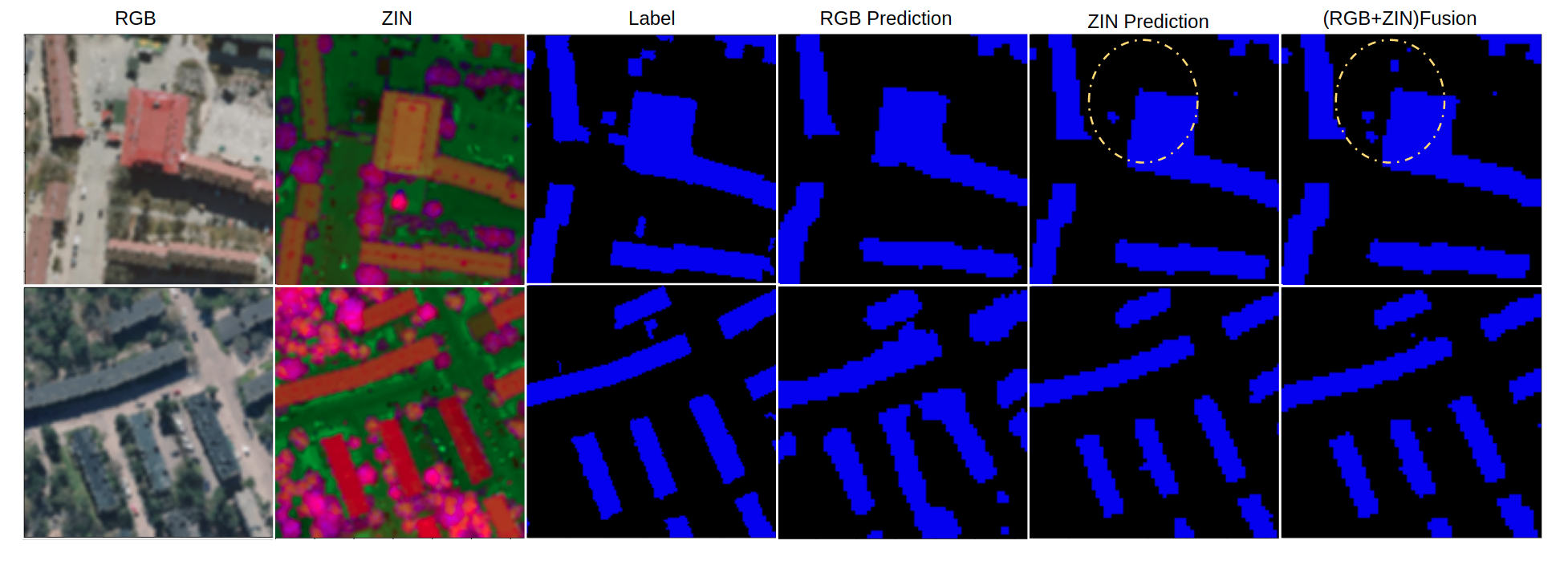}
\end{subfigure}
\vspace*{-3mm}
\caption{Segmentation Result Samples. From left to right, we have RGB image, ZIN 3-channel representation, ground truth, prediction from  the Single Stream Segmentation Network on RGB input, prediction from the Single Stream Segmentation Network on ZIN input and prediction from the Dual Stream Fusion Segmentation Network.}
\label{fig7}
\end{figure*}

These segmentation networks are trained on input patches of size 128x128. The LiDAR samples in our data are of different dimensions. These samples are divided into small patches of size 128x128 and fed into the network with a batch size of 4. We also applied a list of data augmentation to the input patches. Some of the augmentations are horizontal, vertical flips, rotations, crop, pad, translate. The model is trained for 150 epochs using binary cross-entropy loss and Adam optimizer. The learning rate is set to 0.001 along with the decay rate of 0.1 after every 20 steady epochs. These hyperparameter settings are applied for all experiments. We implemented everything in python using Keras framework and the training is conducted on a single Google colab GPU.

\subsection{Change Detection}
\label{sec:cd_method}
For change detection, a 3D matrix with 2-channels is prepared. The first channel is the segmentation prediction from the trained model. The second channel contains elevation values corresponding to the points which are predicted as buildings by our model. This 3D matrix data is prepared for both the 2017 and 2019 samples covering the same area. Now, the channel-wise difference is calculated between the two-year 3D matrix. This difference is then categorized into positive and negative changes. 

The difference in the first channel represents the changes in area or the changes from the ground. The positive change here can be the construction of a new building in the year 2019 and this area used to be either empty or under a different class in 2017. The negative changes can be due to the demolishment of a building after recording the first data samples in 2017. The difference in the second channel is only about changes in elevation i.e building getting taller or shorter. This covers the scenarios, where a particular area was classified as building in 2017 as well as in 2019. But, in one of the years, it was a construction site and in another year the building height has increased by some floors or decreased(i.e while reconstructing the top floor, partially damaged in some disaster event).

Morphological operations are applied to reduce the noise in the change results. The operations used are 'closing' and 'opening' with the kernel size 3. Once the results are refined, the change maps for each category are prepared. Both maps consist of negative and positive changes. For better interpretation, the two change maps are overlaid with the building footprint from 2019. These final change maps are further explained in the result section.

\section{Results and Evaluation}
\label{sec:eval}
First, we present the quantitative and qualitative comparison of segmentation methods described in the method section. The best segmentation method is chosen to conduct the change detection task. In the later section, a qualitative analysis of the change detection results is shown and discussed in detail.

\subsection{Segmentation Evaluation}
\label{sec:seg_eval}
Network performance is assessed in terms of IOU score taking into account True Positives (TP), False Positives (FP) and False Negatives (FN).
The IOU score is described in equation \ref{eq:score}. 

\begin{equation} 
\label{eq:score}
IOU = \frac{TP}{TP+FP+FN}
\end{equation}

\begin{figure*}  
\centering  
\begin{subfigure}
  \centering  
  \includegraphics[trim=1cm 1cm 1cm 1cm, width=\textwidth, height=85mm]{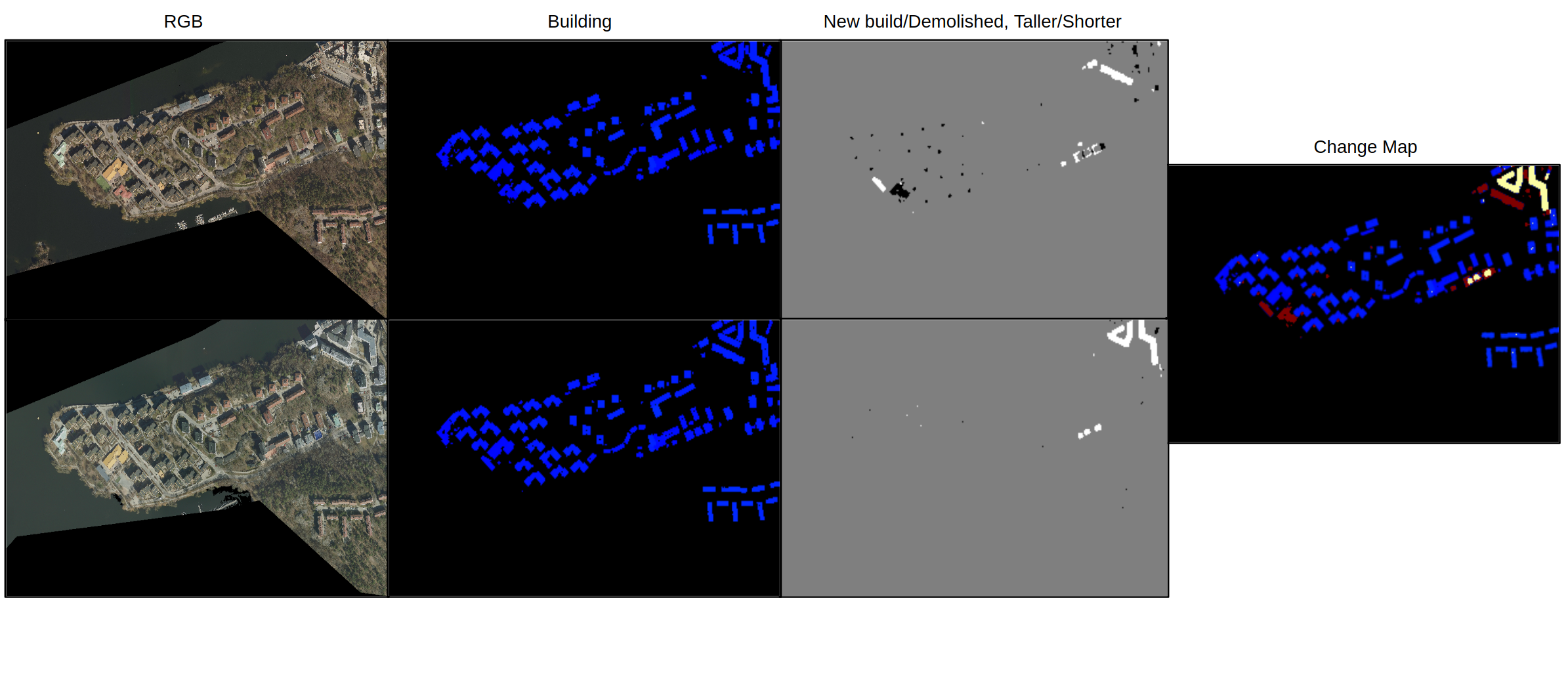}
\end{subfigure}

\begin{subfigure}
  \centering  
  \includegraphics[trim=1cm 1cm 1cm 1cm, width=\textwidth, height=98mm]{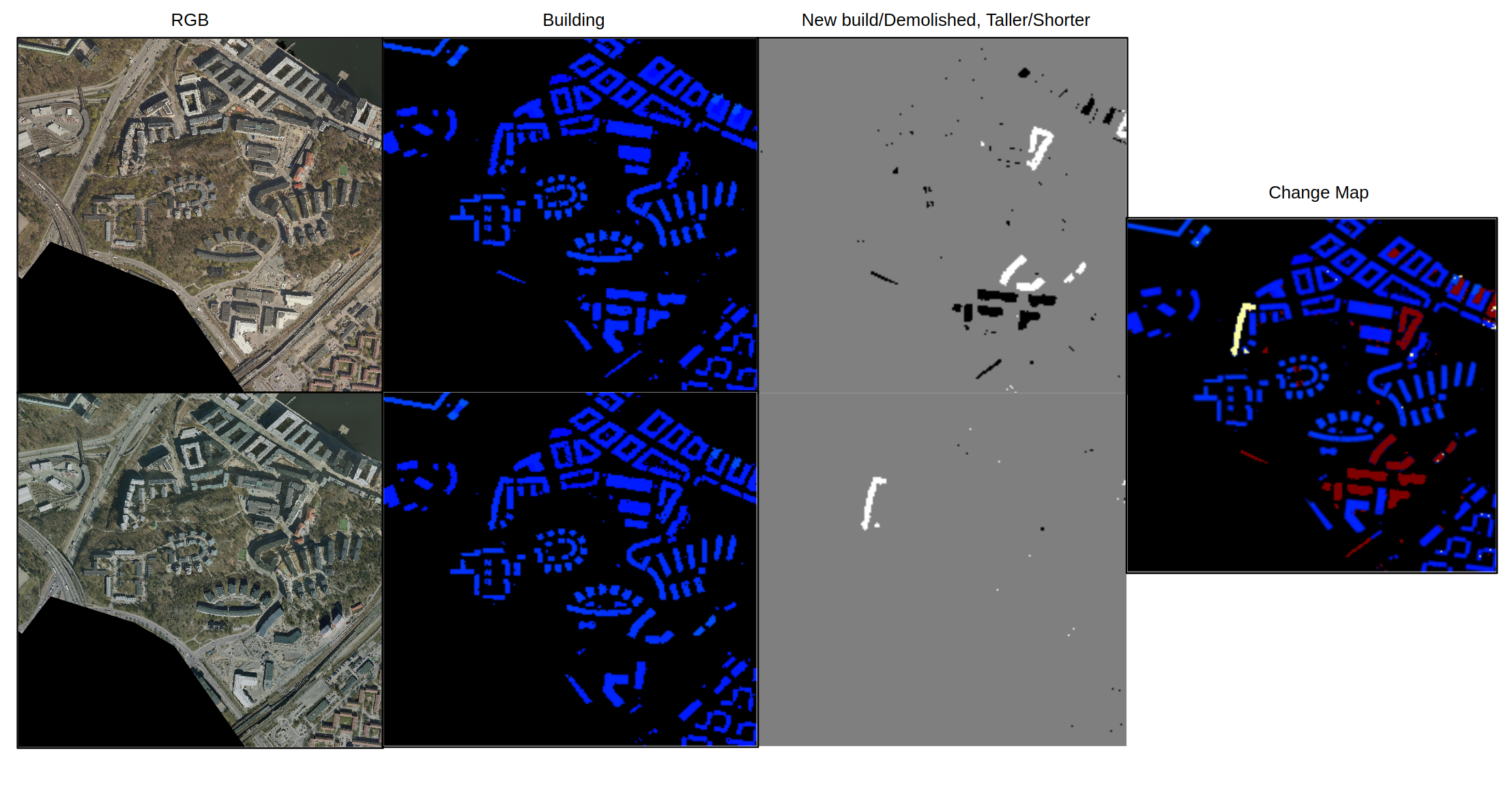}
\end{subfigure}

\caption{Change Detection Result Samples. }
\label{fig4}
\end{figure*}

\begin{figure*}  
\centering  
\begin{subfigure}
  \centering  
  \includegraphics[trim=1cm 0.2cm 1cm 1cm, width=125mm, height=80mm]{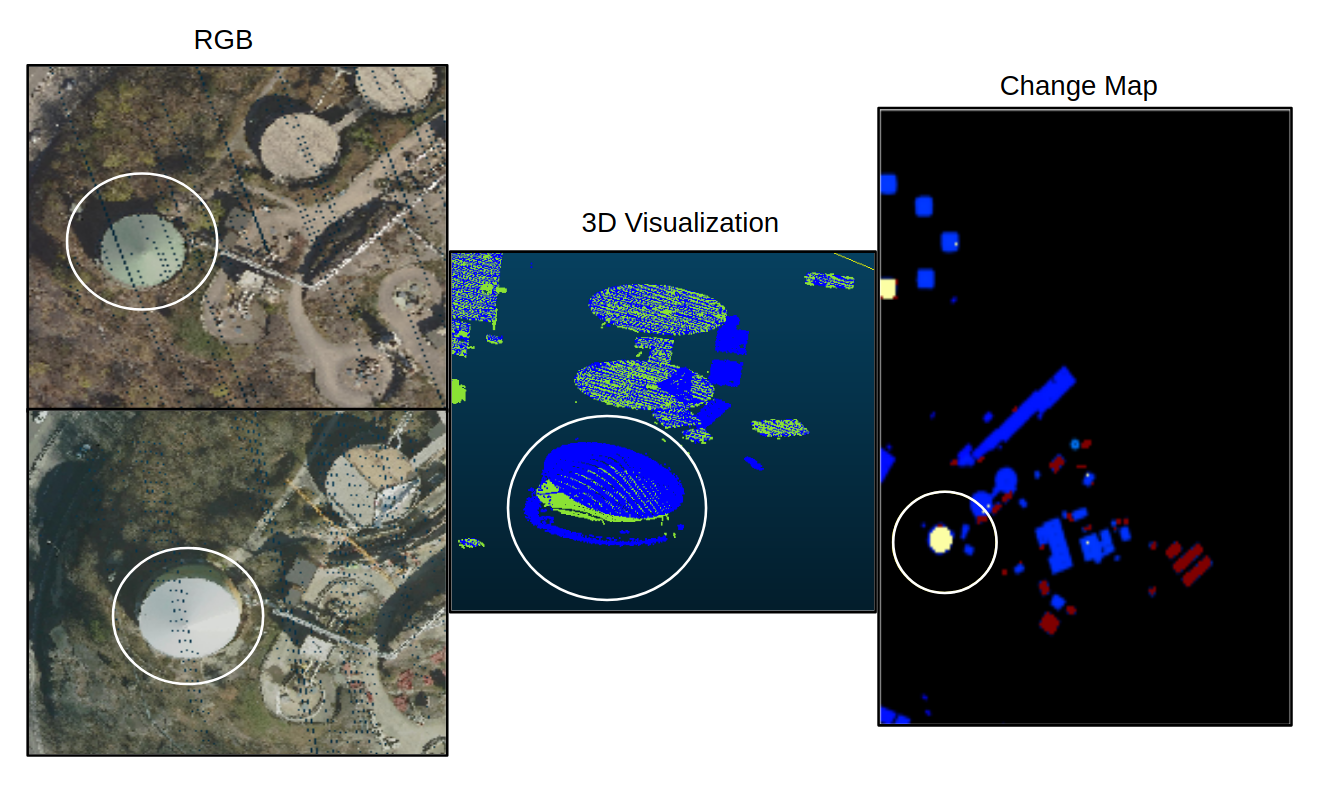}
\end{subfigure}
\caption{Negative Change Detection sample.}
\label{fig3}
\end{figure*}

For the quantitative comparison, Table \ref{SOTA_comp2} list IOU scores of the three methods; single stream segmentation on ZIN, single stream segmentation on RGB and the dual stream fusion network on both RGB and ZIN. The single stream network trained on ZIN input achieved higher IOU score than the one trained on RGB input. The 10\% performance difference is considerably high. The performance is enhanced further by approximately by 2\%, when both RGB and ZIN inputs are fused in training the dual stream fusion network.

The qualitative comparison of the three segmentation methods is visualized in Figure \ref{fig7}. The visualized samples show agreement with the quantitative results and give more insight into specific differences. The performance of the RGB input-based single stream network is substantially low in comparison to others. This can be visually demonstrated by the prediction samples shown in the second row of Figure \ref{fig7}.
In the first row, a small area is highlighted using a dashed circle. According to the label, there are four small buildings with low elevations. The single-stream network detected either one building when trained on ZIN data or none when trained on RGB data. Whereas, the fusion network successfully detected three small buildings. The possible explanation here is that the RGB adds specific building color information to the network. Elevation, intensity, and color information together provide better features to the fusion network, which help in recognizing small buildings. 

Our proposed fusion network scored 86.7\% IOU. We analyzed the remaining 13\% of the building which are not detected by our network. These are the small buildings with low elevations. Network struggles more if these small buildings are partially covered by dense trees. These small buildings or partly not visible in optical images. With current point density the intensity and 'number of returns' also fail to differentiate small buildings from other.

\begin{table}[htbp]
\begin{center}
\resizebox{.6\columnwidth}{!}{%
\begin{tabular}{l|c}
\textbf{Methods} &\textbf{IOU}\\ \hline \hline
{RGB Single Stream} & 75  \\ 
{ZIN Single Stream}  & 85  \\ 
{RGB+ZIN Dual Stream}  & 86.7 \\  
\end{tabular}%
}
\end{center}
\caption{Performance Comparison of Segmentation Methods.}
\label{SOTA_comp2}
\end{table}

\subsection{Change Detection Evaluation}
\label{sec:cd_eval}
Two samples are shown in Figure \ref{fig4} for qualitative analysis of the change detection. The first row of the figure contains the data from the year 2017 and the second row is for 2019 data. From left to right we have: First the RGB visualization of the area, which is not an orthophoto but the RGB information corresponding to the points in 2D surface data. The second is the visualization of the building footprints. The buildings are in blue color and the background is in black. In the third column holds the two change maps. The first change map contains changes corresponding to newly built(positive changes) and demolished buildings(negative changes). The new buildings are in white and demolished buildings are in black. It is crucial to note that this map indicates the area-wise changes along with elevation changes. The second change map contains only elevation changes. These changes exclude all elevation changes pointed by the first change map and focused on increase or decrease in number of floors. The fourth is the visualization of the final change map, where all the changes are projected onto the 2019 building footprint. In the final change map, buildings are in blue, (newly built and demolished buildings) are in red and elevation changes are in off-white.

Following the presented workflow, we have prepared and analyzed the change maps for all the sites in the dataset. The entire process is automatic and can be performed end to end. The detected changes were confirmed by the officials along with comments on false building detections which are are mainly the sheds installed on top of the building during construction. These sheds give an impression of extra height to the building and these elevation differences are falsely detected as building changes. One of such examples is shown in Figure \ref{fig3}. From left to right, we presented RGB visualization of samples from 2017 and 2019 in the first and second row respectively. The second column is the 3D visualization of the point cloud where 2017 data is in Green color and 2019 data is in blue. The problem area is highlighted with a white circle. The third column is the change map, where elevation changes are detected in the highlighted building. The detected change is false positive and happened due to shed installment during the reconstruction of the top floors of the building. It is noteworthy that these false positives are still an indicator of ongoing construction on top of the building and can be utilized for monitoring purposes.

\section{Conclusion}
\label{sec:conclusion}

In this paper, we presented a an approach to handle the irregularity of 3D point cloud data from ALS and utilize efficiently for building change detection. Our method explores the benefit of several LiDAR attributes in 2D which are extracted from the 3D point cloud. We proposed two lightweight networks for segmenting buildings from the background and presented the performance comparison. First architecture operates on single modality either optical or ALS whereas the second architecture benefits from both optical and LiDAR features. 

Our findings indicate that the segmentation results are better when both optical and ALS dat are fused. Our change detection approach generate three change maps explained in the previous section. All change maps are georeferenced and can be visualized in tools like QGIS, ArcGIS, GEE, and CloudCompare. These change maps and quick visualization can help in enhancing the monitoring of construction of new buildings, horizontal extension of building such as illegal balcony extension if visible in top view, demolishment of old buildings, completion/ongoing construction, partial or complete building damage due to disaster events and many others.

{
	\begin{spacing}{1.17}
		\normalsize
		\bibliography{ISPRSguidelines_authors} 
	\end{spacing}
}

\end{document}